\documentclass[10pt,conference,a4paper]{IEEEtran}
\usepackage{graphicx}
\usepackage{xcolor}
\usepackage{subcaption}

\newcommand{\hide}[1]{}

\newcommand{\etal}{\textit{et al.}~}

\ifCLASSINFOpdf
\else
\fi
\usepackage{amsmath}
\usepackage{amssymb}
\usepackage{url}

\begin{document}
%

\title{A CNN-RNN Framework for Image Annotation from Visual Cues and Social Network Metadata}

\author{\IEEEauthorblockN{Tobia Tesan}
\IEEEauthorblockA{Quantexa Ltd, London, UK\\
\tt\small{tobiatesan@quantexa.com}}
\and
\IEEEauthorblockN{Pasquale Coscia}
\IEEEauthorblockA{University of Padova, Italy\\
\tt\small{pasquale.coscia@unipd.it}}
\and
\IEEEauthorblockN{Lamberto Ballan}
\IEEEauthorblockA{University of Padova, Italy\\
\tt\small{lamberto.ballan@unipd.it}}
}



%


\maketitle

\begin{abstract}
Images represent a commonly used form of visual communication among people. Nevertheless, image classification may be a challenging task when dealing with unclear or non-common images needing more context to be correctly annotated. Metadata accompanying images on social-media represent an ideal source of additional information for retrieving proper neighborhoods easing image annotation task. To this end, we blend visual features extracted from neighbors and their metadata to jointly leverage context and visual cues. Our models use multiple semantic embeddings to achieve the dual objective of being robust to vocabulary changes between train and test sets and decoupling the architecture from the low-level metadata representation. Convolutional and recurrent neural networks (CNNs-RNNs) are jointly adopted to infer similarity among neighbors and query images. We perform comprehensive experiments on the NUS-WIDE dataset showing that our models outperform state-of-the-art architectures based on images and metadata, and decrease both sensory and semantic gaps to better annotate images.
\end{abstract}


%
\IEEEpeerreviewmaketitle

\section{Introduction}
\begin{figure}[!t]
\centering
\includegraphics[width=6cm]{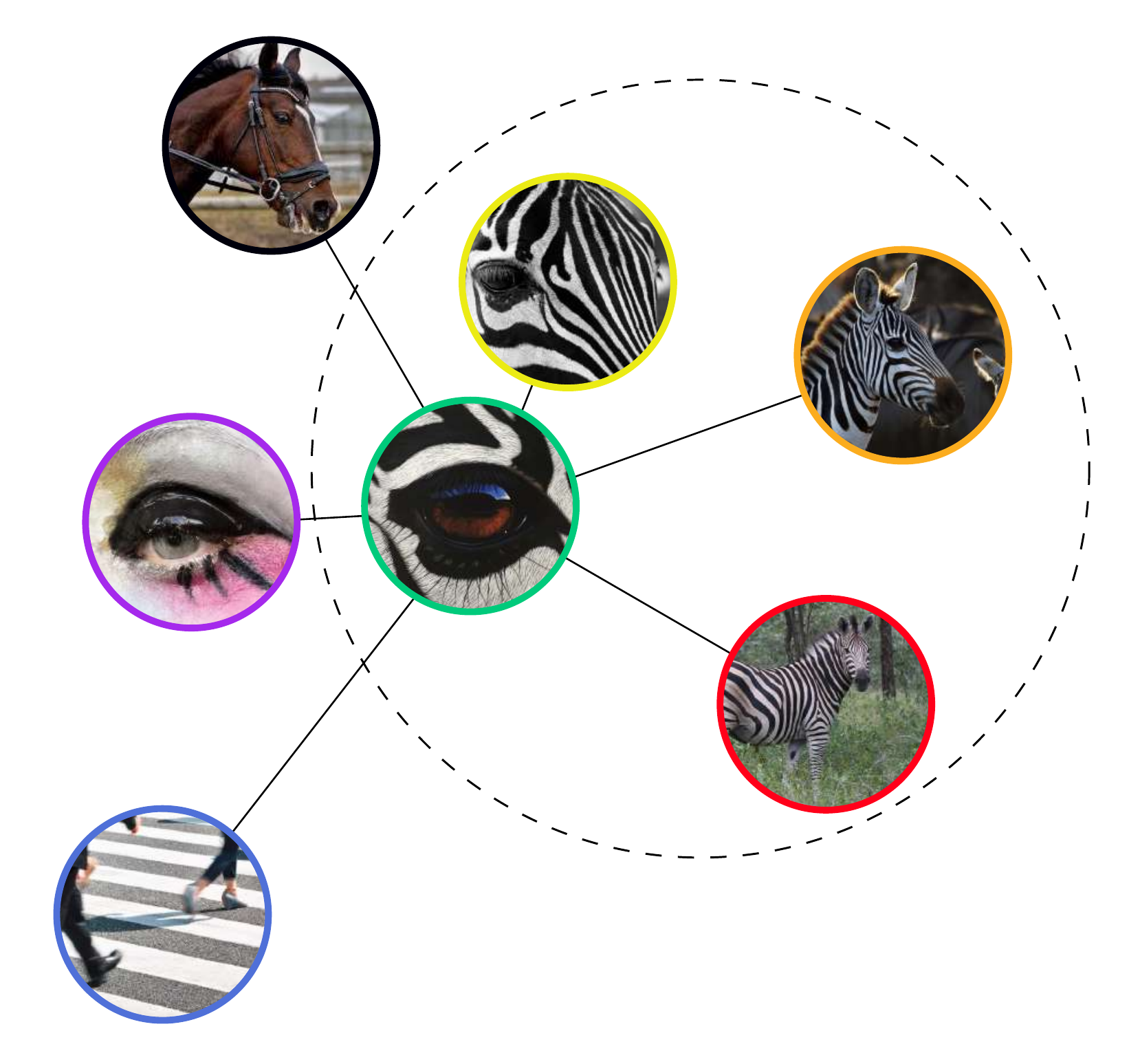}
\caption{Some images might be hard to recognize without additional context. However, related images on a social network typically share similar \textit{metadata}. Based on this intuition, given an image, we retrieve a neighborhood of images sharing similar metadata (e.g. tags) to assist the image annotation task. Our approach builds on \cite{Johnson2015LoveTN} and introduces more advanced semantic mapping and CNN-RNN fusion schemes.\vspace{-7pt}}
\label{fig:intro}
\end{figure}

Images represent an effective and immediate form of expression commonly used to share events and moments of our daily lives. This is particularly true nowadays with the rising popularity of social networks such as Facebook, Twitter and Instagram.
Additional information like similar images and social network \emph{metadata}, are often employed to provide external context and to emphasize moods and messages. Dealing with such contextual data could advantage visual recognition tasks, such as image tagging and retrieval \cite{csur16}, in ambiguous cases where main parts are occluded or unrecognizable (as in Figure \ref{fig:intro}).
In this paper we build on the intuition that a context of additional weakly-annotated images can help in disambiguating the visual classification task, as shown in the seminal work by Johnson \etal \cite{Johnson2015LoveTN}.

The idea of using contextual data to improve visual recognition is not new \cite{Torralba2003,Dvornik2018}. Even humans usually benefit from the context in object detection and scene recognition \cite{Oliva2007}.
In particular, in this work we exploit the (noisy) contextual information given by metadata embedded in images shared on social-networks. Metadata could be very useful to classify examples that occur very rarely or showing visual elements in non-prototypical views. Here image and network metadata can be considerably effective in bridging the sensory and the semantic gap~\cite{Davis2004,McAuley2012}.

Various types of metadata are shared on social-networks. For example, digital photos normally provide information like ISO, exposure, location or timestamp. Users may also add textual descriptions, or provide names of people which appear in photos. Several works have exploited metadata to improve image classification and retrieval, mostly using user-generated tags \cite{Guilla2009,Hwang2012,Gong2013,Niu2014}, GPS data \cite{Hays2008,Tang2015} or groups \cite{Gwang2012}.
In \cite{Johnson2015LoveTN}, image metadata such as tags or Flickr groups are used nonparametrically to generate a pool of related images, that can be further exploited by a deep neural network to blend visual information from a given image and its neighborhood. The key contribution of the approach is a model that can deal with different metadata and adapts over time with no (or very limited) re-training. Thus the model reported state-of-the-art results on multilabel image annotation by taking advantage of strong visual models \cite{Alex2012,Gong2014} and flexible nonparametric approaches \cite{Verma2012,Yu2014}.

In this work we explore different architectures based on both visual cues and external data (e.g., tags) to improve the simple fusion scheme presented in \cite{Johnson2015LoveTN}. 
More specifically, we first focus on preserving distance between a test image $x$ and its neighbors to capture more relevant labels, as well as on handling vocabulary changes when new terms are included.
To this end, our proposed architectures attempt to better encode the semantic meaning of tags through word embeddings \cite{w2v,Saedi2018}.
Second, we investigate and design different architectures for image-to-neighborhood features fusion. Here the main source of inspiration is given by recent CNN-RNN models for image classification and captioning \cite{Wang2016,Liu2017}.
In these works, a CNN is used to extract the image feature vector, which is then fed into an RNN that either decodes it into a list of labels (multilabel image classification) or a sequence of words composing a sentence (captioning).
In contrast, we investigate different strategies in which an RNN is used to sequentially blend the visual or multimodal information in a joint feature space.

The remainder of the paper is organized as follows. In Section \ref{sec:background}, we review related work in the area of image classification in a (noisy) multimodal scenario. In Section \ref{sec:model}, we present our deep network framework.
We evaluate the performance of our method on the NUS-WIDE dataset~\cite{Chua}, and Section \ref{sec:experiments} shows that the approach improves previous state-of-the-art models \cite{Johnson2015LoveTN,Wang2016}.

\section{Related Work}
\label{sec:background}
\subsection{Image Tagging and Retrieval}
The idea of harvesting images from the web to train visual classification models has been explored many times in the past \cite{feifei-2010,xchen-2015,Rupprecht2017}.
Despite its simplicity, a popular and quite effective approach for automatic image annotation, that has been often used in early works, is nearest-neighbors based label transfer \cite{makadia-2008,Verma2012}.
More recently, deep networks have been applied extensively also in this domain achieving state-of-the-art results on many popular benchmarks \cite{Alex2012,Gong2014}.

Among the vast literature on image tagging and retrieval \cite{csur16}, our work is mostly related to multimodal representation learning of images and labels.
To this end, early works often model the association between visual data and labels in a generative way or rely on mapping images and labels to a common semantic space using techniques such as CCA or KCCA \cite{Lavrenko2003,Hwang2012,Uricchio2017}.
Hu \etal~\cite{Hu2015LearningSI} observe that diverse levels of visual categorization are possible depending on the level of desired abstraction. Thus, they rely on structured inference to capture relationships among concepts in neural networks.
In general, these approaches demonstrate the benefit of exploiting side information and correlations between visual features and labels, but they only rely on ground truth annotations.  

\subsection{Automatic Image Annotation with Metadata}
Several previous works tackled the automatic image annotation task using social-network metadata \cite{Davis2004,Hays2008,McAuley2012,Gwang2012}.
User-generated tags are significantly the most commonly used metadata for multilabel image classification. In \cite{Guilla2010}, Guillaumin \etal consider a scenario in which only visual data is used at test time, but metadata from social media websites (such as Flickr) are available at training time and can be leveraged to improve classification using semi-supervised learning. Moreover, a combination of simple nonparametric models and metric learning is used in \cite{Guilla2009}, while \cite{Yu2014} focuses on selecting a better set of training images to drive the label transfer.
Flickr groups are exploited in \cite{Gwang2012} to derive a measure of image similarity which can encode broader correlations than user-generated tags and labels. A graph over tags, groups or common GPS location is used by Niu \etal \cite{Niu2014} to define a semi-supervised topic model for image classification.

Our work falls in this area. Inspired by the model presented by Johnson \etal \cite{Johnson2015LoveTN}, we also use a deep network to blend the visual information extracted from a neighborhood of images sharing similar metadata.
This idea has been also recently followed in~\cite{Zhang_2019_CVPR} where a co-attention mechanism is used to construct a graph in which each node represents a relevant neighbor and correlated images are connected by edges.
Our method differs from these works because we focus on defining a more effective architecture to combine visual cues and social-network metadata from both the test image and the neighborhood.

\section{Our Framework}
\label{sec:model}
\begin{figure*}
    \begin{subfigure}[b]{0.5\textwidth}
    \centering
        \includegraphics[width=7.5cm]{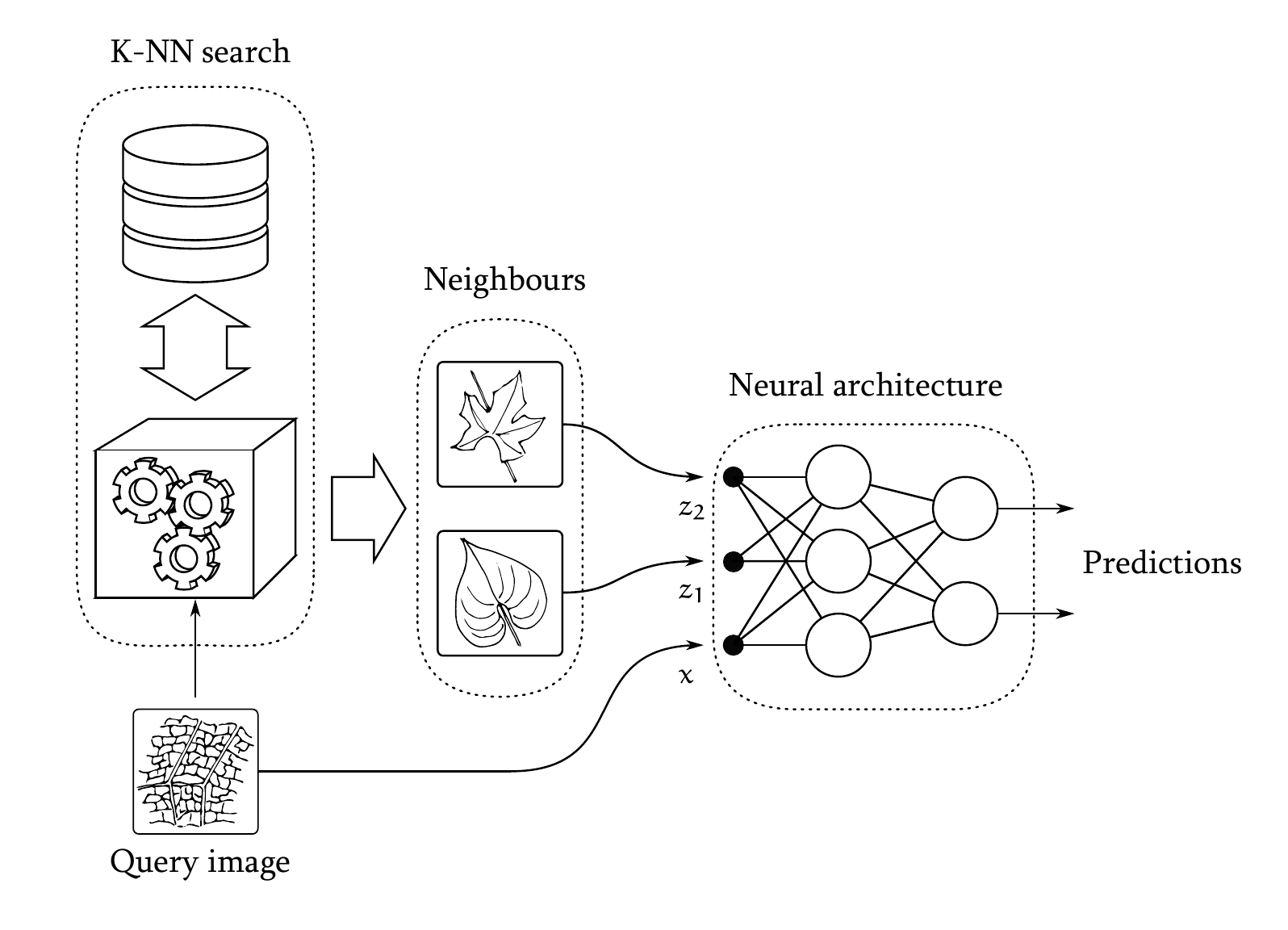}
        \caption{Visual models}
        \label{fig:architecture_visual}
    \end{subfigure}
    ~ 
    \begin{subfigure}[b]{0.5\textwidth}
    \centering
        \includegraphics[width=7.5cm]{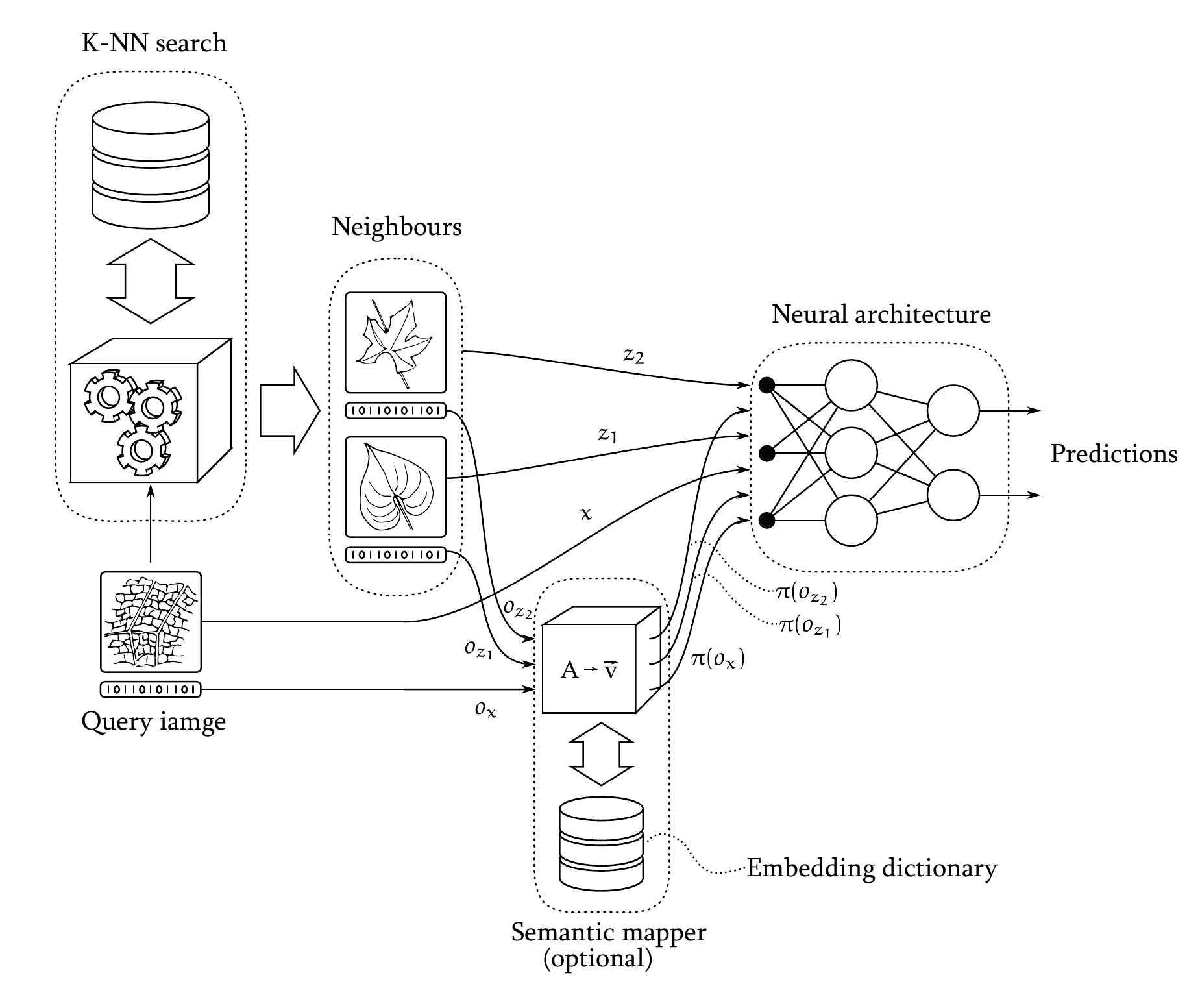}
        \caption{Joint models}
        \label{fig:architecture_joint}
    \end{subfigure}
    ~ 
    \vspace{-7pt}
    \caption{General architectures of the proposed models. K-NN is used to retrieve similar images using metadata, while a neural network processes the retrieved information. (a) shows the architecture for visual models (as in~\cite{Johnson2015LoveTN}) where only visual features for both query image and neighbors are used to predict labels. (b) shows the architecture for joint models where metadata are also fed, possibly after a transformation step, to the final classification layer.\vspace{-7pt}}
    \label{fig:frameworks}
\end{figure*}

Our goal is to annotate images using side information carried by their neighbors. More specifically, we jointly exploit visual features as well as tags which commonly accompany images on social networks. Tags are embedded using different semantic mappings. Our models are built upon the work presented by Johnson \textit{et al.}~\cite{Johnson2015LoveTN}, where metadata are only used to retrieve similar images and the annotation task mainly relies on visual features.
We propose two general architectures for images annotation, both based on visual features and image metadata (see Figure~\ref{fig:frameworks}). 
Whereas visual models only exploit visual cues, joint models handle metadata which are directly fed to the neural network after a transformation step. 

All the models generate nonparametrically a neighborhood $Z_x$ for a query image $x$ using metadata and then the networks are trained to classify $x$ given its neighbors in $Z_x$. The neighborhood generation process is parametrized over a neighborhood size $m$ and a max rank $M$. More specifically, let $\mathcal{Z}_x$ be the $M-$nearest neighbors of $x$ according to a distance measure $\delta$. The set of candidate neighborhoods for an image $x$ is the set:
\begin{equation}
    Z_x = \{s \in \mathcal{P}(X): |s| = m\},
\end{equation}
where $\mathcal{P}(X)$ denotes the power set of $X$, that is the set of considered images.
The prediction $s(x, \theta)$ is the average of $f(x, \vec{z}; \theta)$ over all candidate neighborhoods:
\begin{equation}
    s(x, \theta) = \frac{1}{|Z_x|}\sum_{z \in Z_x} f(x, \vec{z}; \theta),
\end{equation}
where $x$ is the image to be classified, $\vec{z} = (z_1, z_2, ..., z_m)$ are the neighbors and $f(x, \vec{z}; \theta)$ is the output of the neural network which takes into account their visual cues. 

The model is trained by computing a loss function $\mathcal{L}$ and minimizing: 
\begin{equation}
    \theta^{*} = \arg\min_{\theta} \sum_{(x, y) \in D_{train}} \mathcal{L}(s(x, \theta), y),
\end{equation}
where $y$ represent a subset of all possible labels that appear in $D$.
Note that neighbors are ordered according to their distance when fed to the neural network and thus the network may learn to treat the closest ones differently.

Joint models differ from visual models in that they enrich image representation with additional information. More specifically, such models use metadata which are directly fed to the final layer of the network after a transformation step $\pi(\cdot)$ involving a lookup in a dictionary of semantic embeddings.
In this case, the prediction $s(x, \theta)$ is the average of $f(x, \pi(o_x), \vec{z}, \pi(\vec{o_z}); \theta)$, where $o_x$ is the metadata vector for image $x$ while $\pi(o_x)$ is its transform. 
We shall use $\pi(\vec{o}_z)$ as shorthand for map $(\pi, {\vec{o}}_z) = (\pi(o_{z_1}), \pi(o_{z_2}), ..., \pi(o_{z_m}))$, where $\vec{o}_z$ are metadata vectors for the neighborhood. 

\subsection{Metadata Encoding}
Metadata representation may affect network's ability to recover correct annotations. For this reason, we firstly encode metadata without associating any meaningful representation to each word, i.e., semantically close words could be associated to distant vectors, and secondly consider more powerful word encoding techniques.
\smallskip

\subsubsection{One-hot Encoding}
We focus on social-network tags represented as binary vectors $o_x \in \{0, 1\}^\tau$. More specifically, let $x$ the query image and $(t_{(1)}, t_{(2)}, ..., t_{(n)})$ all relevant tags for $x$ chosen from a vocabulary of $\tau$ tags, the binary vector $o_x$ is the sum of the one-hot vectors for each of its tags:
\begin{equation}
        o_x = \sum_{i~s.t.~t_i \in \{t_{(1)}, t_{(2)}, ..., t_{(n)}\}} e^{\tau}_i.
\end{equation}
Using \texttt{id}, i.e., raw binary vectors, neighborhoods are computed using the Jaccard distance $\mathcal{J}$ between binary vectors.
Binary vectors $o_x$ for each image $x$ (or neighbor $z_i$) are directly handled by the neural network, without further processing. The Jaccard distance is defined as follows:
\begin{equation}
    \mathcal{J}(x, x^{'}) = 1 - \frac{|t_x \cap t_{x^{'}}|}{|t_x \cup t_{x^{'}}|}
\end{equation}
with $\mathcal{J}(x, x) = 0$.
\smallskip

\subsubsection{Semantic-aware Encoding}
We also explore more powerful word embedding techniques in order to encode similar word into similar vectors. We consider a transformation that maps a vector $o_x$ to a \textit{semantic space} $\pi:\{0, 1\}^{\tau} \rightarrow \mathbb{R}^n$. It is clear that, unlike visual models, where metadata are used implicitly, a neural network trained to make predictions as a function of one or more binary vectors becomes useless if the vocabulary changes. Semantic maps $\pi$ can decouple the low-level bit representation from the semantic meaning, making models learned on a tag vocabulary applicable to a different one, as long as an appropriate $\widetilde{\pi}$ is available that maps the \textit{new} binary vectors onto the \textit{old} semantic space. More specifically, given a map or dictionary of embeddings $\beta:TAGS\rightarrow\mathbb{R}^n$ for some $n$, we define $\rho(o_x; \beta)$ as the sum of the vectors $\beta(t_{(i)})$ for each tag $t_{(i)}$ relevant for image $x$, i.e.:
\begin{equation}
    \rho(o_x; \beta) = \sum_{i = 1}^\tau o_{x_{(i)}}\cdot\beta({t_{(i)}}). 
    \label{eq:weight}
\end{equation}

For $\pi(x) = \rho(o_x; \beta)$, we consider two semantic embeddings. Firstly, we use a dictionary of \texttt{word2vec} embeddings~\cite{w2v}; they are obtained by training on a $100$-billion-words subset of the Google News database and contain $300$-dimensional vectors for $3$ million words and phrases. We expect to recover some semantic information from the tags and improve performance, as well as achieving decoupling from the low-level binary representation for joint architectures. We choose cosine distance for $\delta$, defined as: 
\begin{equation}
    sim_{cos}(x_1, x_2) = 1-\frac{\vec{x_1}\cdot\vec{x_2}}{|\vec{x_1}||\vec{x_2}|}.
\end{equation}

Secondly, we use \texttt{WordNet} embeddings which works in the same fashion as \texttt{word2vec}, except that $\beta$ is extracted from a dictionary where vector representations are optimized to be similar if the words are close on the WordNet taxonomy.
Cosine distance is again our choice for $\delta$.
\texttt{WordNet} embeddings~\cite{Saedi2018} comprise a dictionary of $650$-dimensional vectors obtained from Princeton WordNet $3.0$\footnote{\url{https://github.com/nlx-group/WordNetEmbeddings}} with $60,000$ words.

\subsection{Visual Models}

Visual models only rely on extracted visual features of input images without considering additional information. We consider three visual models based on fully-connected and recurrent layers.

\begin{figure*}
    \begin{subfigure}[b]{0.24\textwidth}
        \includegraphics[width=\linewidth]{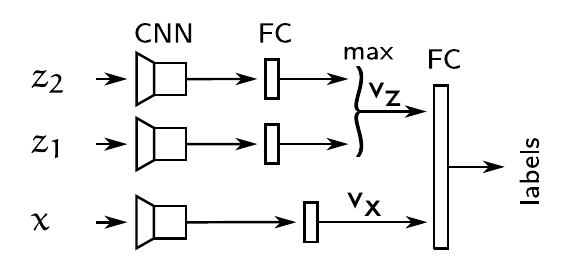}
        \caption{\texttt{LTN}}
        \label{fig:visual1}
    \end{subfigure}
    \begin{subfigure}[b]{0.24\textwidth}
        \includegraphics[width=\linewidth]{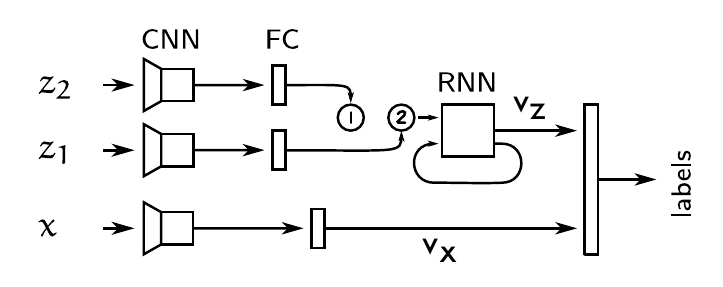}
        \caption{\texttt{RTN}}
        \label{fig:visual2}
    \end{subfigure}
    \begin{subfigure}[b]{0.24\textwidth}
        \includegraphics[width=\linewidth]{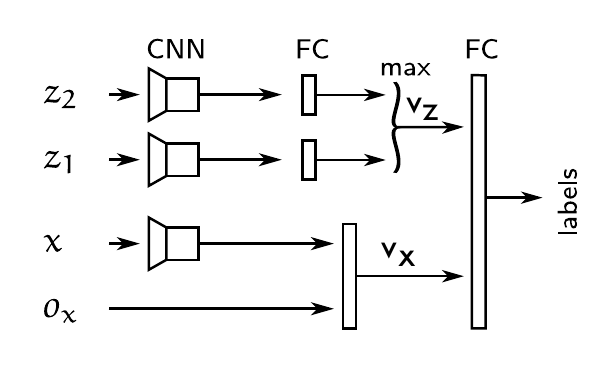}
        \caption{\texttt{LTN+Vecs}}
        \label{fig:joint1}
    \end{subfigure}
    \begin{subfigure}[b]{0.24\textwidth}
        \includegraphics[width=\linewidth]{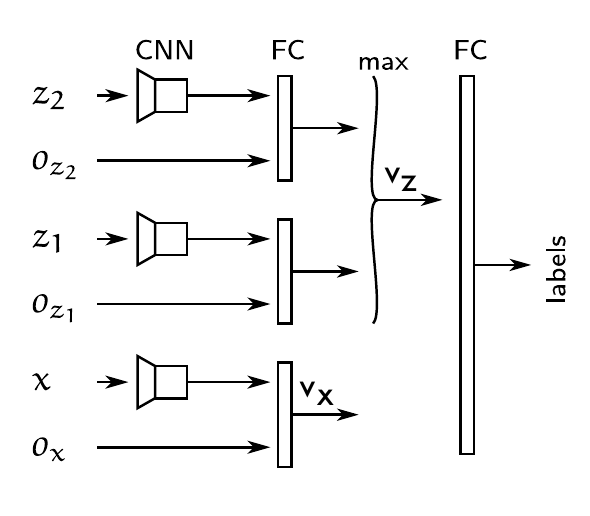}
        \caption{\texttt{LTN+AllVecs}}
        \label{fig:joint2}
    \end{subfigure}
    
    \begin{subfigure}[b]{0.24\textwidth}
        \includegraphics[width=\linewidth]{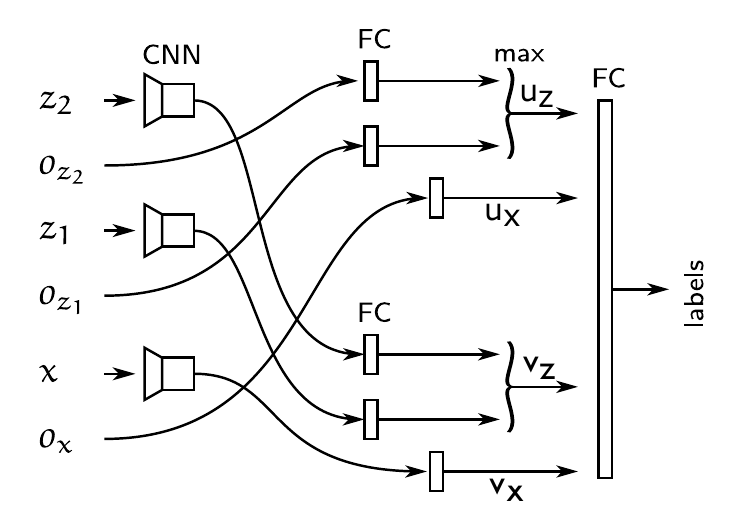}
        \caption{\texttt{LTwin}}
        \label{fig:joint3}
    \end{subfigure}
    \begin{subfigure}[b]{0.24\textwidth}
        \includegraphics[width=\linewidth]{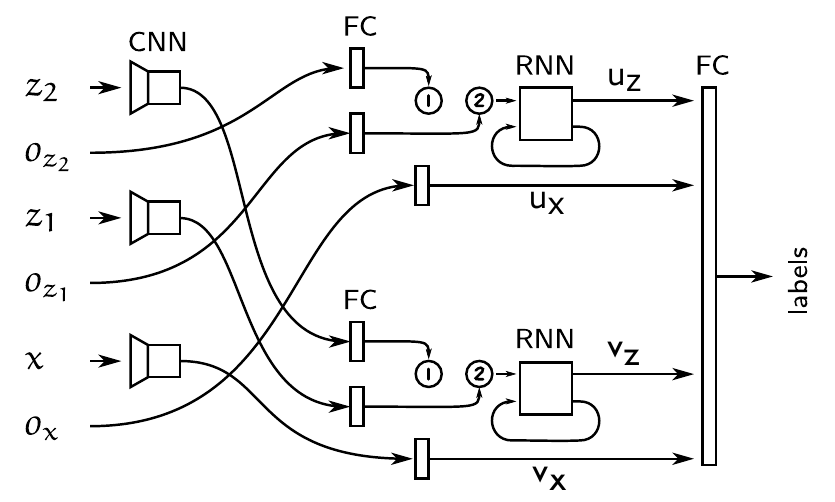}
        \caption{\texttt{LTwin+RNN}}
        \label{fig:joint4}
    \end{subfigure}
    \begin{subfigure}[b]{0.24\textwidth}
    \includegraphics[width=\linewidth]{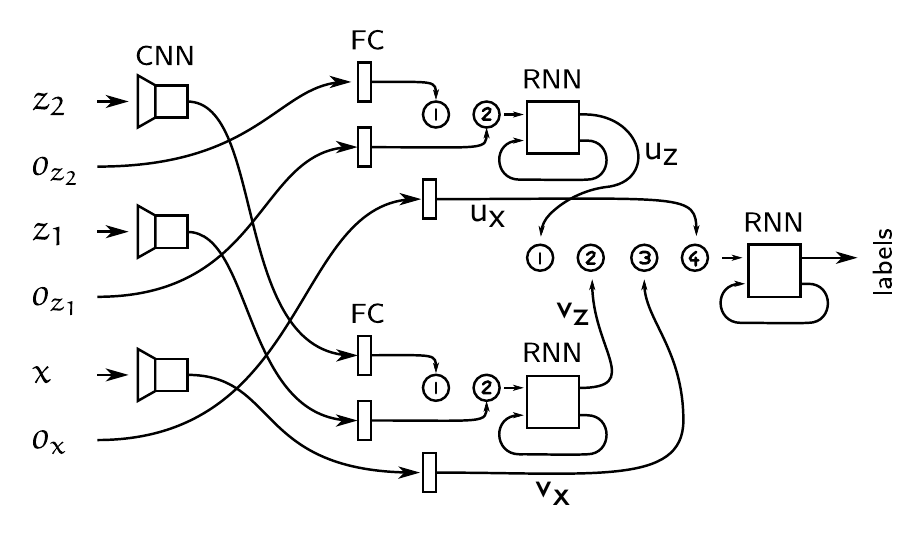}
        \caption{\texttt{LTwin+2RNN}}
        \label{fig:joint5}
    \end{subfigure}
    \begin{subfigure}[b]{0.24\textwidth}
    \includegraphics[width=\linewidth]{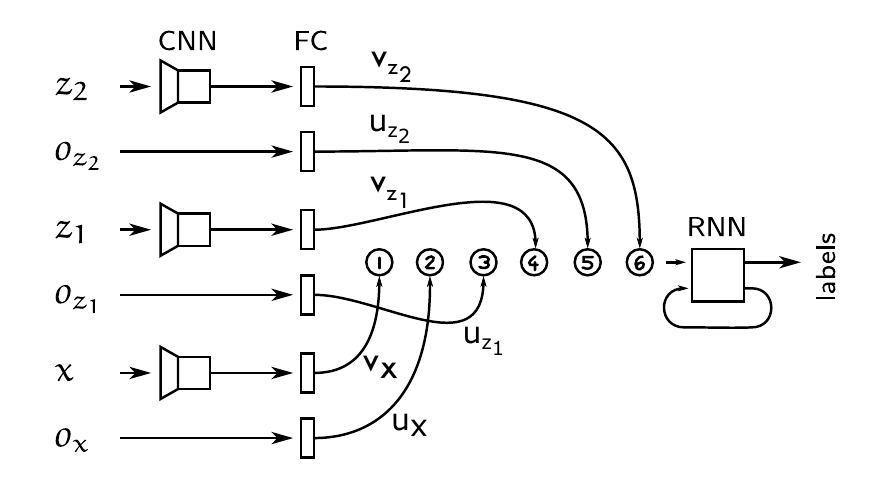}
        \caption{\texttt{LZip}}
        \label{fig:joint6}
    \end{subfigure}
    \caption{Our architectures which leverage image features along with metadata. (a),(b) represent two visual models where metadata are not employed. (c)--(h) are different ways to fuse image features and metadata. In this work, as metadata we only use tags and we exploit recurrent layers and semantic embeddings in order to leverage contextual information.\vspace{-2pt}}
\end{figure*}

\subsubsection{Visual-only}
This architecture acts as baseline; it simply amounts to a fully-connected layer over visual features $\phi(x)$ output by a CNN for an image $x$. Therefore, 
\begin{equation}
    f(x, \vec{z};\theta) = W_y\Phi(x)+b_y
\end{equation} 
Note that $\vec{z}$ is not used.

\subsubsection{\texttt{LTN}}
This is the model proposed in \cite{Johnson2015LoveTN}. The label scores are computed as follows:
\begin{equation}
    f(x, \vec{z}; \theta) = W_y \begin{bmatrix} v_x\\v_z \end{bmatrix} +b_y 
\end{equation}
where $\vec{z} = (z_1, z_2, ..., z_m)$ is a vector of neighbors obtained nonparametrically, $x$ is the image to be classified, and 
\begin{equation}
    v_x = \sigma(W_x\Phi(x) + b_x),
\end{equation}
\begin{equation}
    v_z = \max_{i = 1, ..., m} (\sigma(W_z\Phi(z_i) + b_z))
\end{equation}
where $\sigma$ is a ReLU activation function. 
The model is depicted in Figure~\ref{fig:visual1}. Note that the weights $W_z$ and $b_z$ are shared among all $(z_1, z_2, ..., z_m)$ and $v_x, v_z \in \mathbb{R}^h$.

\subsubsection{\texttt{RTN}}
This architecture extends \texttt{LTN} by replacing the max-pooling operation with a RNN in order to better discriminate individual neighbors.
Sequential image processing may allow the network to retain only relevant features of the neighborhood handling each image separately.

More specifically, the hidden
state $v_z$ is defined as follows:
\begin{equation}
    v_z = RNN((z_1, z_2, ..., z_m); W_{RNN}),
\end{equation}
where the notation $RNN((i_1, ..., i_n), W)$ denotes a recurrent neural network sequentially fed with inputs $(i_1, ..., i_n)$ while $W$ are the corresponding parameters. In this case, RNN is a long short-term memory (LSTM) network with linear activation function. The other parameters remain unchanged. 
The model is depicted in Figure~\ref{fig:visual2}.

\subsection{Joint Models}
Joint models are directly fed with metadata instead of leveraging metadata only implicitly along with visual features. Metadata improve the semantic level detected by extracted visual features. In the following, we define several architectures handling metadata (or their embeddings) using linear and recurrent layers.

\subsubsection{\texttt{LTN+Vecs}}
This architecture makes use of metadata $o_x$, i.e., metadata of image to be classified, which are concatenated to the output of the CNN of image $x$. 

The output of the network is defined as follows:
\begin{equation}
    f(x, \pi(o_x), \vec{z}; \theta) = W_y \begin{bmatrix} v_x \\ v_z \end{bmatrix} + b_y,
\end{equation}

where 
\begin{equation}
    v_x = \sigma \begin{pmatrix} W_x \begin{bmatrix} \Phi(x) \\ \pi(o_x)\end{bmatrix} + b_x\end{pmatrix}.
\end{equation}
$v_z$ is defined as in \texttt{LTN} visual model.
Note that such model does not use neighbor metadata vectors and it only relies on visual features of closest images. A transformation step is then applied to map metadata onto a new space (see Figure~\ref{fig:joint1}).

\subsubsection{\texttt{LTN+AllVecs}}
This architecture, unlike the previous one, uses metadata vectors $o_x$ of the image to be classified and metadata of its neighbors $\vec{o}_z$. 

The output is defined as follows:
\begin{equation}
    f(x, \pi(o_x), \vec{z}, \pi(\vec{o}_z); \theta) = W_y \begin{bmatrix} v_x \\ v_z \end{bmatrix} + b_y,
\end{equation}
where $v_x$ is defined as above and
\begin{equation}
    v_z = \max_{i = 1, ..., m} \sigma \begin{pmatrix} W_z \begin{bmatrix} \Phi(z_i)\\\pi(o_{z_i}) \end{bmatrix}\end{pmatrix}.
\end{equation}
In this case, $\sigma$ is a ReLU activation function. The model is depicted in Figure~\ref{fig:joint2}.

\subsubsection{\texttt{LTwin}}
Unlike \texttt{LTN+AllVecs}, such architecture processes features and metadata using two separate pipelines, i.e., metadata are not concatenated with the images features. The neighbors are blended with a max-pooling layer, so the model is not able to discriminate between nearest and farthest neighbors.

The output of the network is defined as follows:
\begin{equation}
    f(x, \pi(o_x), \vec{z}, \pi(\vec{o_z}); \theta) = W_y \begin{bmatrix} v_x\\v_z\\u_x\\u_z\end{bmatrix} + b_y,
\end{equation} 
where $v_x$ and $v_z$ are defined as in the \texttt{LTN} model, while $u_x = \sigma(W_{x_u}\pi(o_x)+b_{x_u})$ and $u_z = \max_{i = 1, ..., m}~\sigma(W_{z_u}\pi(o_{z_i})+b_{z_u})$. 
Max-pooling is applied on both neighbors' features and their metadata. The model is depicted in Figure~\ref{fig:joint3}.

\subsubsection{\texttt{LTwin+RNN}}
Unlike the previous architecture, such model replaces max-pooling layers with RNN networks to handle the neighbors. Once again, RNN is an LSTM with linear activation. 
The output is equal to \texttt{LTwin} architecture with $v_z = RNN((FC_{z_1}, ..., FC_{z_m}); W_{RNN})$ and $u_z = RNN((FC_{o_{z_1}}, ..., FC_{o_{z_m}}); W_{o_{RNN}})$, where $FC_{(\cdot)}$ are outputs of fully-connected layers applied to image features and metadata, respectively. 
The model is depicted in Figure~\ref{fig:joint4}.

\subsubsection{\texttt{LTwin+2RNN}}
This architecture differs from the previous one in that the final fully connected layer is also replaced with a RNN. The output is defined as follows:
\begin{equation}
    f(x, \pi(o_x), \vec{z}, \pi(\vec{o_z}); \theta) = RNN((v_x, v_z, u_x, u_z); W_{f_{RNN}}),
\end{equation}
where $v_x,v_z, u_x$ and $u_z$ are defined as in \texttt{LTwin+RNN}. 
The model is depicted in Figure~\ref{fig:joint5}.

\subsubsection{\texttt{LZip}}
Finally, this architecture uses just one RNN to combine features and metadata which are separately processed by FC layers. The output is defined as follows:
\begin{multline}
    f(x, \pi(o_x), \vec{z}, \pi(\vec{o_z}); \theta) = \\RNN((v_x, u_x, v_{z_1}, u_{z_1}, ..., v_{z_m}, u_{z_m}); W_{RNN}).
\end{multline}    
The model is depicted in Figure~\ref{fig:joint6}.

\subsection{Implementation Details}
We use RMSProp algorithm with He-Zhang initialization~\cite{He2015} and apply dropout with $p = 0.5$. We also set batch size dimension to $64$ (in lieu of $50$, as found in \cite{Johnson2015LoveTN}) and $h = 500$.
We apply $L_2$ regularization with $\lambda = 3 \times 10^{-4}$ and use a learning rate of $1 \times 10^{-4}$. $\lambda$ was chosen with grid search. 
We use early stopping with a maximum of $10$ and a minimum of $3$ epochs, incremented to $15$ and $5$ for joint models, respectively. 
We run experiments with $(3, 6), (6, 12)$ and $(12, 24)$ as choices of $(m,M)$. Our CNN is the ImageNet pre-trained AlexNet~\cite{Alex2012} model available on Caffe, as in~\cite{Johnson2015LoveTN}.

\section{Experiments}
\label{sec:experiments}
\subsubsection{Dataset}
We use the NUS-WIDE dataset~\cite{Chua} which comprises $269,648$ images uploaded on the photo sharing website Flickr, annotated with $81$ ground truth labels for evaluation. 
NUS-WIDE is highly unbalanced over classes, whereas the tag \texttt{sky} is relevant for around $53,000$ images, many classes have less than a thousand images. 
We restrict ourselves to the fixed subset of $190,253$ images used in~\cite{Johnson2015LoveTN,Zhang_2019_CVPR} for ease of comparison.
The dataset comprises $422,364$ unique Flickr tags, which we narrow down to the $\tau = 5000$ most frequent tags. The dataset is randomly partitioned to form training, validation and test sets of $110,000$, $40,000$ and $40,253$ images, respectively. We average the results over $5$ of such splits.

\subsubsection{Metrics}
We report per-label and per-image mean Average Precision (mAP), as well as precision and recall. Note that, in this area, the most common evaluation protocol assumes that an algorithm should assign a fixed number $k$ of labels to each image. To this end, following prior work \cite{Gong2014,Johnson2015LoveTN,Wang2016}, we report results for $k = 3$. Since on NUS-WIDE the average number of labels per image is approx. $2.4$, by assigning exactly $3$ labels, no classifier can achieve unit precision and recall (thus we report on Table~\ref{table:comparison} the real upper bound for each metric).
However, as also highlighted in \cite{Guilla2009,Johnson2015LoveTN,csur16}, mAP directly measures ranking quality, so it naturally handles multiple labels and does not require to set a fixed number $k$. Therefore, mAP is the primary evaluation metric used further on in our evaluation.

\subsection{Experimental Results}
Table~\ref{table:comparison} shows our best results in comparison to several baselines and state-of-the-art models. 
Firs of all, the \texttt{LTwin} model outperforms the other methods on both mAP metrics. It is also important to  note that for the corresponding models proposed in~\cite{Johnson2015LoveTN}, our implementation of \texttt{LTN} achieves comparable results while \texttt{LTN+Vecs} has worse performance. 
Therefore, the \texttt{LTwin} model achieves best results showing a $10$ and $2$ percentage performance increase on both mAP metrics w.r.t. the corresponding \texttt{LTN+Vecs} baseline.

\begin{table*}[!th]
\centering
\resizebox{\textwidth}{!}{\begin{tabular}{l||c|c|cc|cc } 
Method & mAP$_{lab}$ & mAP$_{img}$ & rec$_{lab}$ & prec$_{lab}$ & rec$_{img}$ & prec$_{img}$\\\hline\hline
Tag-only Model + linear SVM~\cite{McAuley2012}& 46.67 & - & - & - & - & -\\
Graphical Model (all metadata)~\cite{McAuley2012} & 49.00 & - & - & - & - & -\\
CNN + WARP~\cite{Gong2014} & - & - & 35.60 & 31.65 & 60.49 & 48.59\\
CNN-RNN~\cite{Wang2016} & - & - & 30.40 & 40.50 & 61.70 & 49.90\\
SR-RNN~\cite{Liu2017} & - & - & 50.17~$\star$ & 55.65~$\star$ & 71.35~$\star$ & 70.57~$\star$\\
SR-RNN + Vecs~\cite{Liu2017}~$\dagger$ & - & - & 58.52~$\star$ & 63.51~$\star$ & 77.33~$\star$ & 76.21~$\star$\\
SRN~\cite{Zhu2017} & 60.00 & 80.60 & 41.50~$\star$ & 70.40~$\star$ & 58.70~$\star$ & 81.10~$\star$\\
MangoNet~\cite{Zhang_2019_CVPR} & 62.80 & 80.80 & 41.00~$\star$ & 73.90~$\star$ & 59.90~$\star$ & 80.60~$\star$\\
LTN~\cite{Johnson2015LoveTN} & 52.78 $\scriptstyle{\pm 0.34}$ & 80.34 $\scriptstyle{\pm 0.07}$ & 43.61 $\scriptstyle{\pm 0.47}$ & 46.98 $\scriptstyle{\pm 1.01}$ & 74.72 $\scriptstyle{\pm 0.16}$ & 53.69 $\scriptstyle{\pm 0.13}$\\
LTN + Vecs~\cite{Johnson2015LoveTN}~$\dagger$ & 61.88 $\scriptstyle{\pm 0.36}$ & 80.27 $\scriptstyle{\pm 0.08}$ & 57.30 $\scriptstyle{\pm 0.44}$ & 54.74 $\scriptstyle{\pm 0.63}$ & 75.10 $\scriptstyle{\pm 0.20}$ & 53.46 $\scriptstyle{\pm 0.09}$\\
\hline\hline
Upper bound & 100.00 $\scriptstyle{\pm 0.00}$ & 100.00 $\scriptstyle{\pm 0.00}$ & 65.82 $\scriptstyle{\pm 0.35}$ & 60.68 $\scriptstyle{\pm 1.32}$ & 92.09 $\scriptstyle{\pm 0.10}$ & 66.83 $\scriptstyle{\pm 0.12}$\\
Our baseline: v-only & 45.05 $\scriptstyle{\pm 0.11}$ & 76.88 $\scriptstyle{\pm 0.11}$ & 42.31 $\scriptstyle{\pm 0.59}$ & 43.74 $\scriptstyle{\pm 1.07}$ & 71.41 $\scriptstyle{\pm 0.13}$ & 51.36 $\scriptstyle{\pm 0.13}$\\
Our baseline: LTN$_{\texttt{n:id}}$ & 53.17 $\scriptstyle{\pm 0.12}$ & 79.82 $\scriptstyle{\pm 0.16}$ & 45.67 $\scriptstyle{\pm 1.75}$ & 47.64 $\scriptstyle{\pm 2.18}$ & 74.29 $\scriptstyle{\pm 0.13}$ & 53.34 $\scriptstyle{\pm 0.17}$\\
Our baseline: LTN + Vecs$_{\texttt{n:id,f:id}}~\dagger$ & 54.86 $\scriptstyle{\pm 0.20}$ & 81.34 $\scriptstyle{\pm 0.15}$ & 46.56 $\scriptstyle{\pm 1.39}$ & 50.10 $\scriptstyle{\pm 1.70}$ & 75.67 $\scriptstyle{\pm 0.17}$ & 54.37 $\scriptstyle{\pm 0.14}$\\
\hline
Our model: RTN$_{\texttt{n:w2v}}$ & 55.36 $\scriptstyle{\pm 0.34}$ & 79.77  $\scriptstyle{\pm 0.27}$ & 48.73 $\scriptstyle{\pm 2.77}$ & 51.21 $\scriptstyle{\pm 2.61}$ & 74.35 $\scriptstyle{\pm 0.29}$ & 53.28 $\scriptstyle{\pm 0.24}$\\
Our model: LTwin$_{\texttt{n:w2v,f:w2v}}~\dagger$ & \textbf{63.13} $\scriptstyle{\pm 0.31}$ & \textbf{83.77} $\scriptstyle{\pm 0.06}$ & 54.40 $\scriptstyle{\pm 1.33}$ & 51.86 $\scriptstyle{\pm 1.58}$ & 78.06 $\scriptstyle{\pm 0.05}$ & 55.78 $\scriptstyle{\pm 0.13}$
\end{tabular}}
\caption{Results on NUS-WIDE. We run on $5$ splits and report mean and standard deviation. Models that also use metadata are marked with $\dagger$. In our models \texttt{n} refers to the encoding used to build the neighborhood, while \texttt{f} to the encoding used to represent image metadata. Models such as \cite{Liu2017} can decide their own prediction length and are not limited by the parameter $k$. In these cases (marked with $\star$) the upper bound does not apply and results are no directly comparable with other approaches.}
\label{table:comparison}
\end{table*}

\begin{table}[ht]
\centering
\begin{tabular}{c||c||c|c } 
Arch & n & mAP$_{lab}$ & mAP$_{img}$\\\hline
\texttt{LTN} & \texttt{id} & 53.17 $\scriptstyle{\pm 0.12}$ & 79.82 $\scriptstyle{\pm 0.16}$\\\hline
\texttt{LTN} & \texttt{w2v} & 54.54 $\scriptstyle{\pm 0.13}$ & \textbf{80.32} $\scriptstyle{\pm 0.16}$\\\hline
\texttt{LTN} & \texttt{wnet} & 53.07 $\scriptstyle{\pm 0.17}$ & 79.95 $\scriptstyle{\pm 0.24}$\\\hline
\texttt{RTN} & \texttt{id} & 53.97 $\scriptstyle{\pm 0.27}$ & 79.23 $\scriptstyle{\pm 0.27}$\\\hline
\texttt{RTN} & \texttt{w2v} & \textbf{55.36} $\scriptstyle{\pm 0.34}$ & 79.77 $\scriptstyle{\pm 0.27}$\\\hline
\texttt{RTN} & \texttt{wnet} & 53.76 $\scriptstyle{\pm 0.33}$ & 79.45 $\scriptstyle{\pm 0.30}$\\
\end{tabular}
\caption{Visual Models results for neighborhood size $(m,M)=(12,24)$. Column \texttt{n} refers to the metadata encoding used to build the neighborhood.\vspace{-2pt}}
\label{table:mM1224visual}
\end{table}

\begin{table}[ht]
\centering
\resizebox{\columnwidth}{!}{%
\begin{tabular}{c||c|c||c|c } 
Arch & n & f & mAP$_{lab}$ & mAP$_{img}$\\\hline
\texttt{LTN+Vecs} & \texttt{id} & \texttt{id} &  54.86 $\scriptstyle{\pm 0.20}$ & 81.34 $\scriptstyle{\pm 0.15}$\\\hline
\texttt{LTN+AllVecs} & \texttt{id} & \texttt{id} &  56.61 $\scriptstyle{\pm 0.12}$ & 81.28 $\scriptstyle{\pm 0.21}$\\\hline
\texttt{LZip} & \texttt{id} & \texttt{id} & 60.64 $\scriptstyle{\pm 0.14}$ & 82.42 $\scriptstyle{\pm 0.32}$\\\hline
\texttt{LZip} & \texttt{w2v} & \texttt{id} & 61.24 $\scriptstyle{\pm 0.51}$ & 82.36 $\scriptstyle{\pm 0.41}$\\\hline
\texttt{LZip} & \texttt{w2v} & \texttt{w2v} & 60.19 $\scriptstyle{\pm 0.57}$ & 82.32 $\scriptstyle{\pm 0.15}$\\\hline
\texttt{LZip} & \texttt{id} & \texttt{w2v} & 62.33 $\scriptstyle{\pm 0.16}$ & 82.91 $\scriptstyle{\pm 0.18}$\\\hline
\texttt{LTwin} & \texttt{id} & \texttt{id} & 56.79 $\scriptstyle{\pm 0.24}$ & 82.64 $\scriptstyle{\pm 0.08}$\\\hline
\texttt{LTwin} & \texttt{id} & \texttt{w2v} & 63.09 $\scriptstyle{\pm 0.16}$ & 83.70 $\scriptstyle{\pm 0.14}$\\\hline
\texttt{LTwin} & \texttt{w2v} & \texttt{id} & 57.73 $\scriptstyle{\pm 0.17}$ & 83.00 $\scriptstyle{\pm 0.06}$\\\hline
\texttt{LTwin} & \texttt{w2v} & \texttt{w2v} & \textbf{63.13} $\scriptstyle{\pm 0.31}$ & \textbf{83.77} $\scriptstyle{\pm 0.06}$\\\hline
\texttt{LTwin} & \texttt{id} & \texttt{wnet} & 55.12 $\scriptstyle{\pm 0.25}$ & 81.48 $\scriptstyle{\pm 0.10}$\\\hline
\texttt{LTwin} & \texttt{wnet} & \texttt{id} & 56.83 $\scriptstyle{\pm 0.24}$ & 82.64 $\scriptstyle{\pm 0.10}$\\\hline
\texttt{LTwin} & \texttt{wnet} & \texttt{wnet} & 54.01 $\scriptstyle{\pm 0.14}$ & 81.06 $\scriptstyle{\pm 0.10}$\\\hline
\texttt{LTwin+RNN} & \texttt{id} & \texttt{id} & 58.87 $\scriptstyle{\pm 0.43}$ & 82.95 $\scriptstyle{\pm 0.08}$\\\hline
\texttt{LTwin+2RNN} & \texttt{id} & \texttt{id} & 62.00 $\scriptstyle{\pm 1.44}$ & 80.52 $\scriptstyle{\pm 2.79}$\\\hline
\texttt{LTwin+2RNN} & \texttt{id} & \texttt{w2v} & 63.04 $\scriptstyle{\pm 0.22}$ & 83.02 $\scriptstyle{\pm 0.34}$\\\hline
\texttt{LTwin+2RNN} & \texttt{w2v} & \texttt{w2v} & 62.33 $\scriptstyle{\pm 0.33}$ & 82.72 $\scriptstyle{\pm 0.37}$\\\hline
\texttt{LTwin+2RNN} & \texttt{id} & \texttt{wnet} & 62.35 $\scriptstyle{\pm 0.56}$ & 82.56 $\scriptstyle{\pm 0.26}$
\end{tabular}
}
\caption{Joint Models results for neighborhood size $(m,M)=(12,24)$,
  and different metadata encodings. Column \texttt{n} refers to the encoding used to build the neighborhood, \texttt{f} to the encoding used as representation: \texttt{w2v} = \texttt{word2vec}, \texttt{wnet} = \texttt{wordnet}, and \texttt{id} refers to raw binary vectors.\vspace{-2pt}}
\label{table:mM1224joint}
\end{table}

\begin{figure}[!t]
\centering
\includegraphics[width=4.1cm,height=3.4cm]{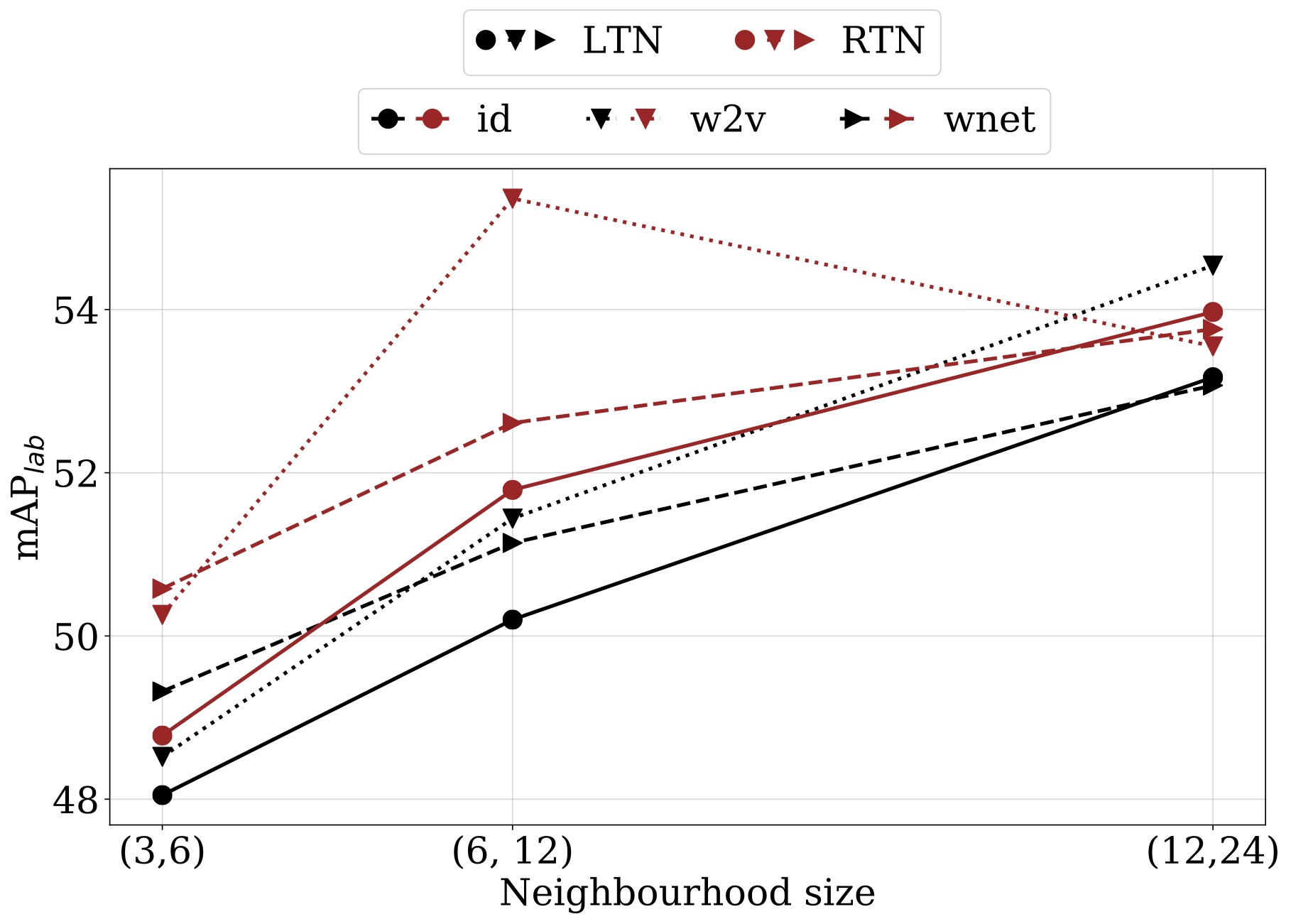}
\includegraphics[width=4.1cm,height=3.4cm]{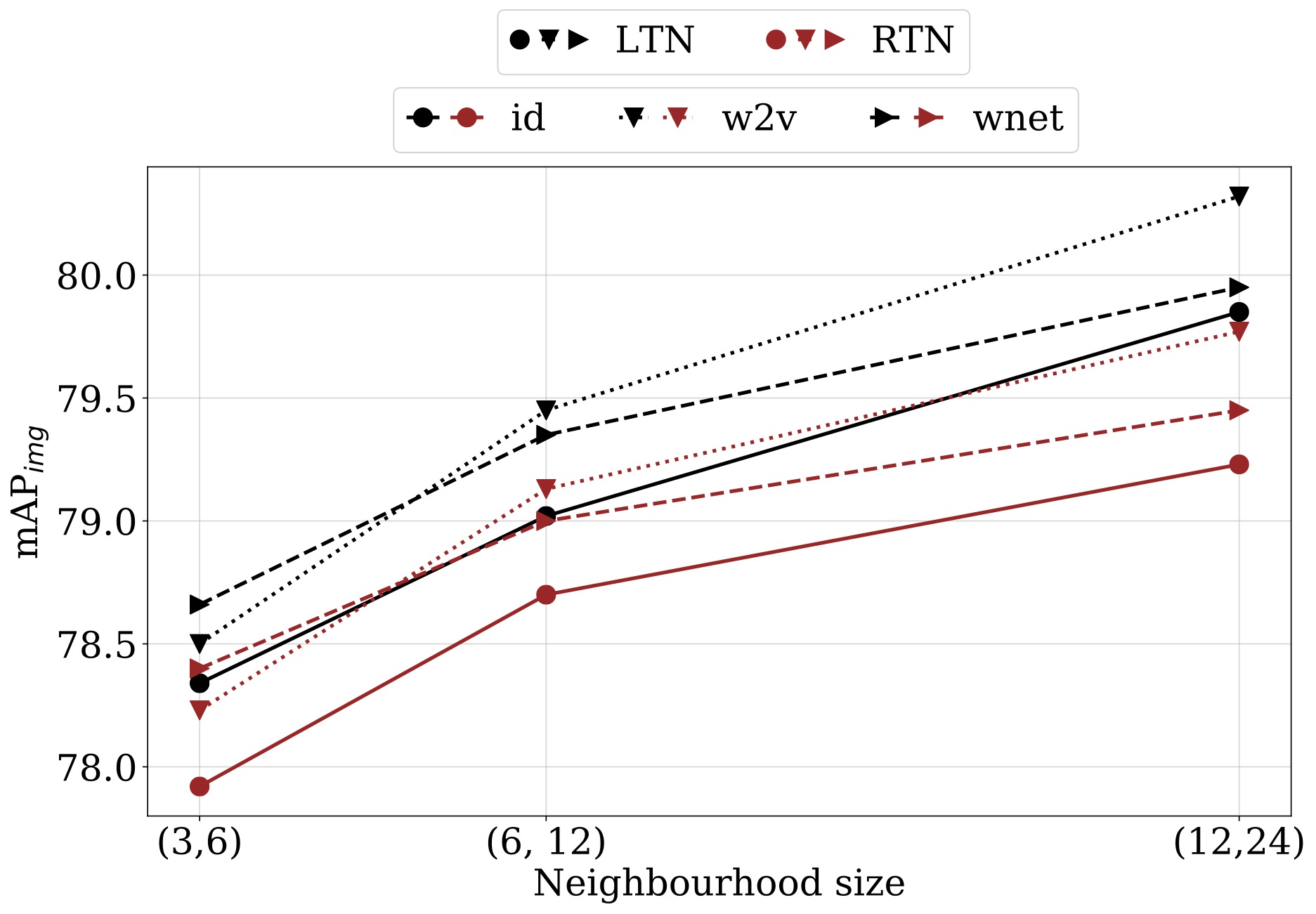}
\caption{mAP$_{lab}$ and mAP$_{img}$ for visual models varying the
  neighborhood size and semantic mapping to retrieve the
  neighbors. Black color refers to \texttt{LTN} model while the red one to \texttt{RTN} model. All the models outperform the visual-only baseline.}
\label{fig:Allvisual}
\end{figure}

\begin{figure}[!t]
\centering
\includegraphics[width=4.1cm,height=3.4cm]{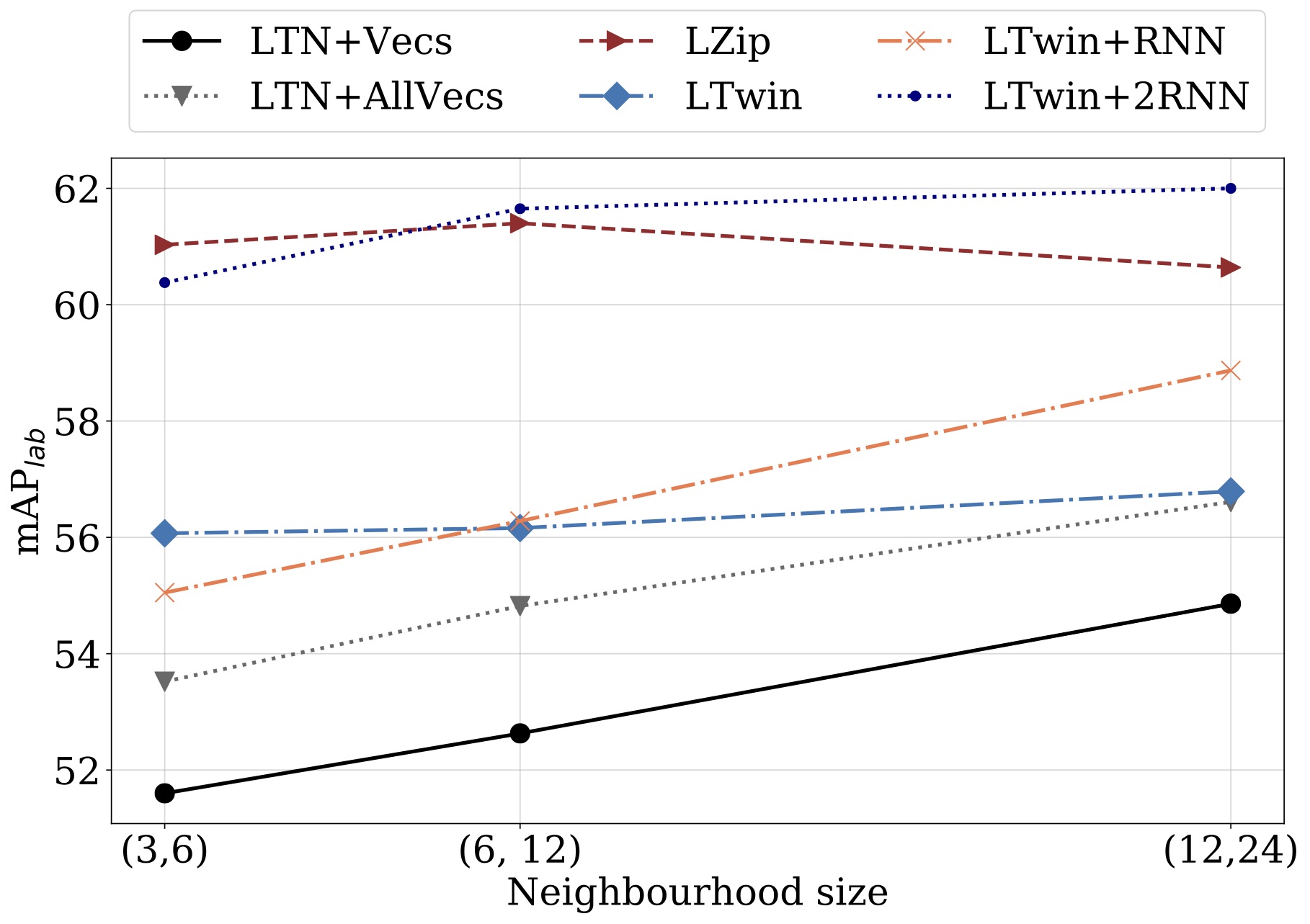}
\includegraphics[width=4.1cm,height=3.4cm]{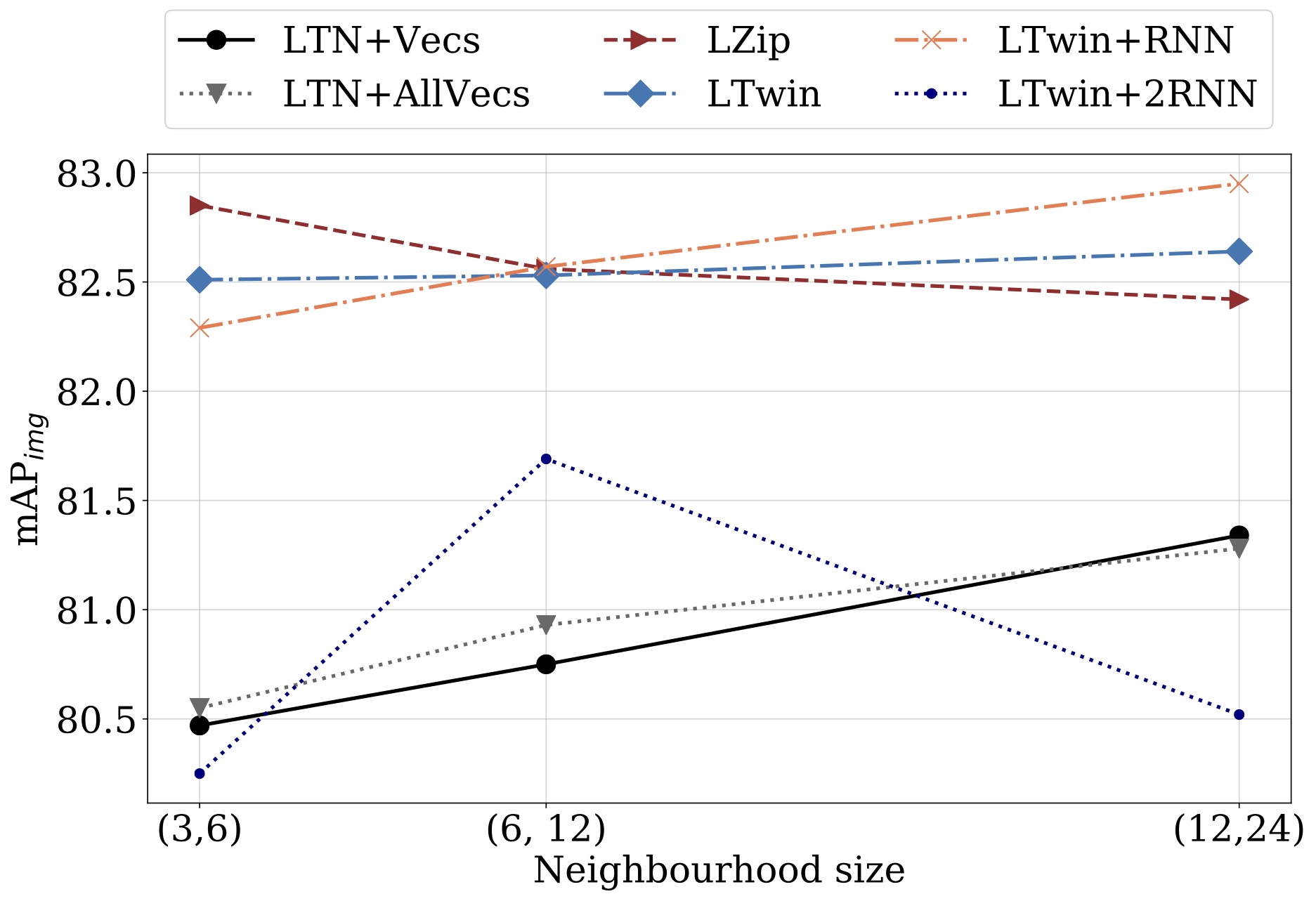}
\caption{mAP$_{lab}$ and mAP$_{img}$ for joint models varying the neighborhood size considering $\pi=$\texttt{id} both for neighbors retrieval and metadata embedding.}
\label{fig:AllJointid}
\end{figure}

More detailed results about all the different architectures presented in Section~\ref{sec:model} are reported in Table~\ref{table:mM1224visual} and Table~\ref{table:mM1224joint} (all the results refer to a neighborhood size of $(12, 24)$, highlighting a vast range of different combinations of architectures and encodings. 
We choose to focus our attention on mAP$_{lab}$ and mAP$_{img}$ since they better summarize classification performances. In general, we note that mAP$_{lab}$ is the metric that is affected the most, whereas mAP$_{img}$ remains more stationary.

\subsubsection{Visual Models}
As shown in Figure~\ref{fig:Allvisual}, for the same neighborhood, \texttt{RTN} leads to an improvement of mAP$_{lab}$ of around $0.7$ to $1.2$ percentage points over \texttt{LTN}, in exchange for a drop of $0.2$ to $0.4$ percentage points of mAP$_{img}$. More interestingly, the gap between $\pi =$  \texttt{id} and \texttt{word2vec} is larger for \texttt{RTN} at low values of $m$. Notice how \texttt{RTN} with \texttt{word2vec} embeddings and a $(3,6)$ neighborhood outperforms “vanilla” \texttt{LTN} with $(6,12)$ neighborhood in terms of mAP$_{lab}$, with negligible impact on mAP$_{img}$. The performance of \texttt{RTN} begins to decline faster than \texttt{LTN} with $\pi =$ \texttt{WordNet}
. This leads to hypothesize that \texttt{RTN} is particularly sensitive to the quality of neighborhoods it is trained on. All models improve monotonically with $m$. 

\begin{figure}[!t]
\centering
\includegraphics[width=4.1cm,height=3.4cm]{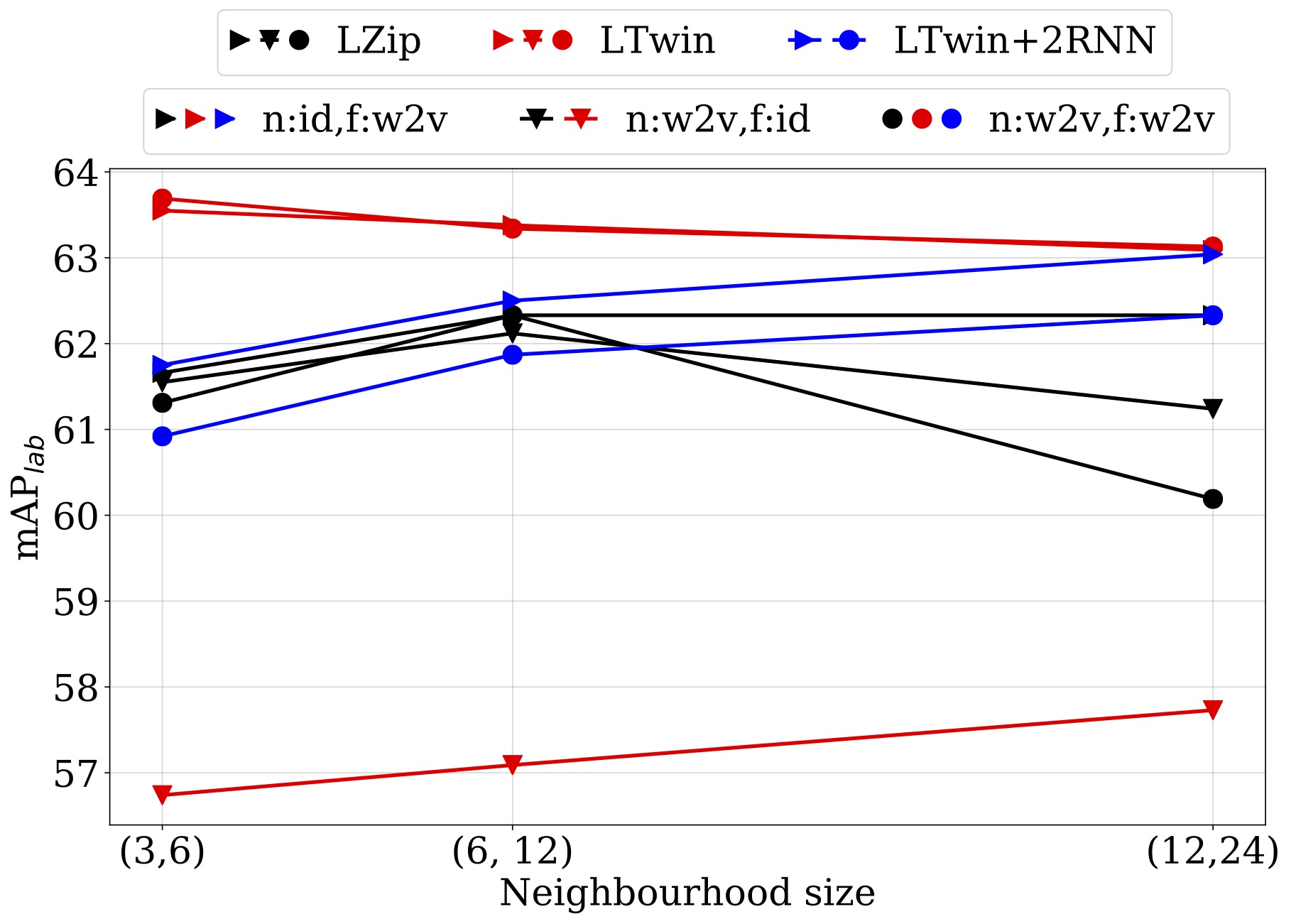}
\includegraphics[width=4.1cm,height=3.4cm]{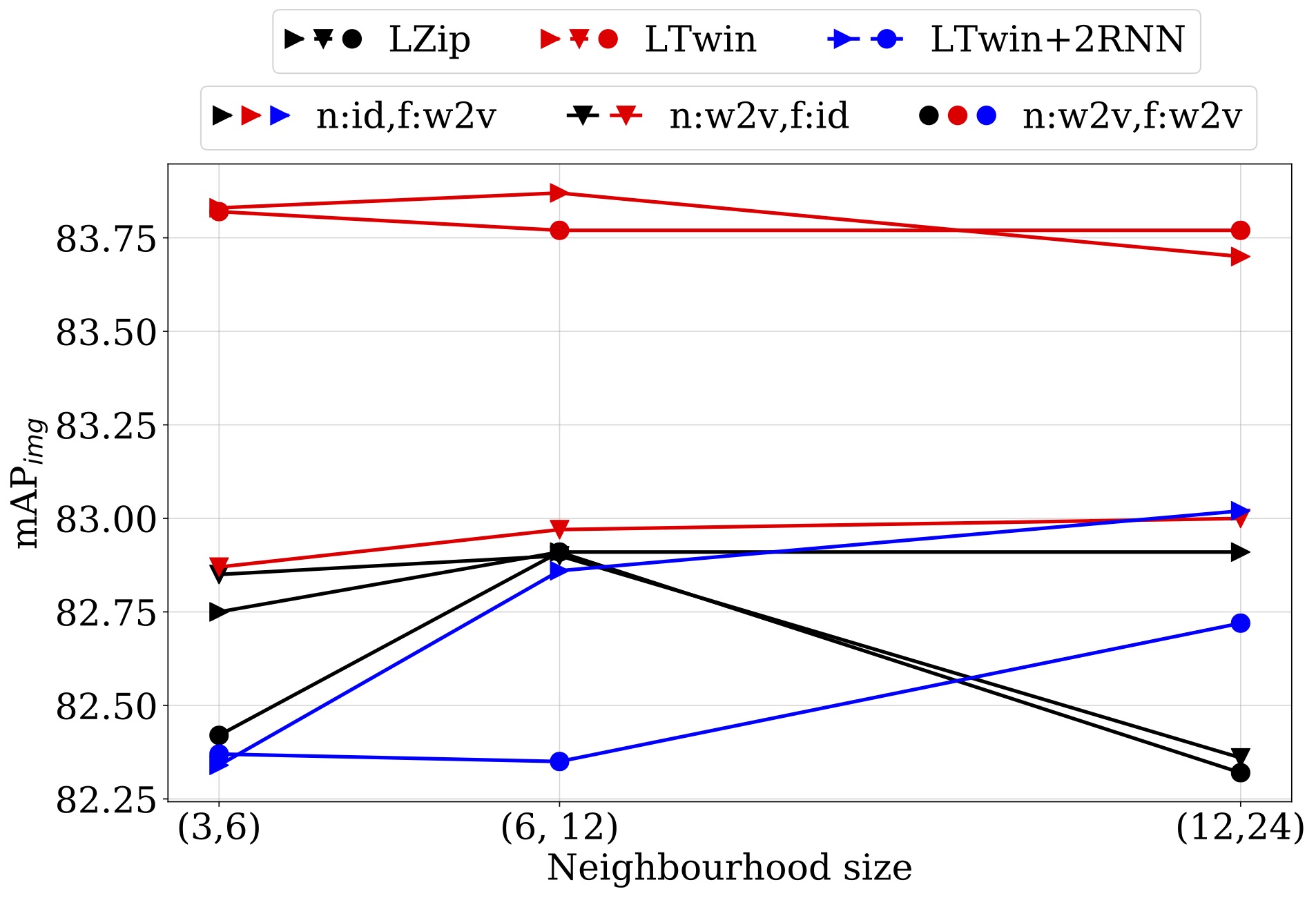}\\
\includegraphics[width=4.1cm,height=3.4cm]{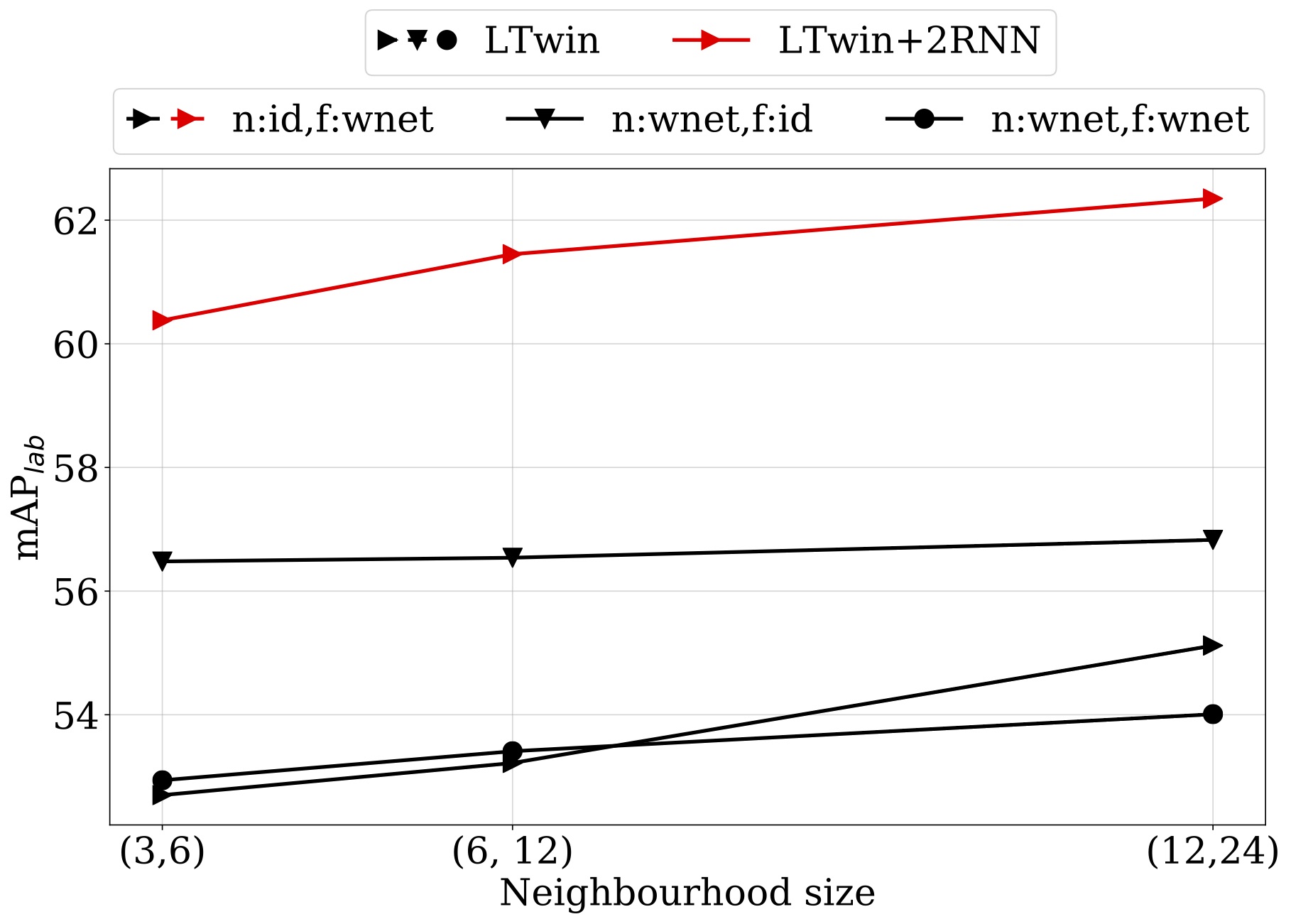}
\includegraphics[width=4.1cm,height=3.4cm]{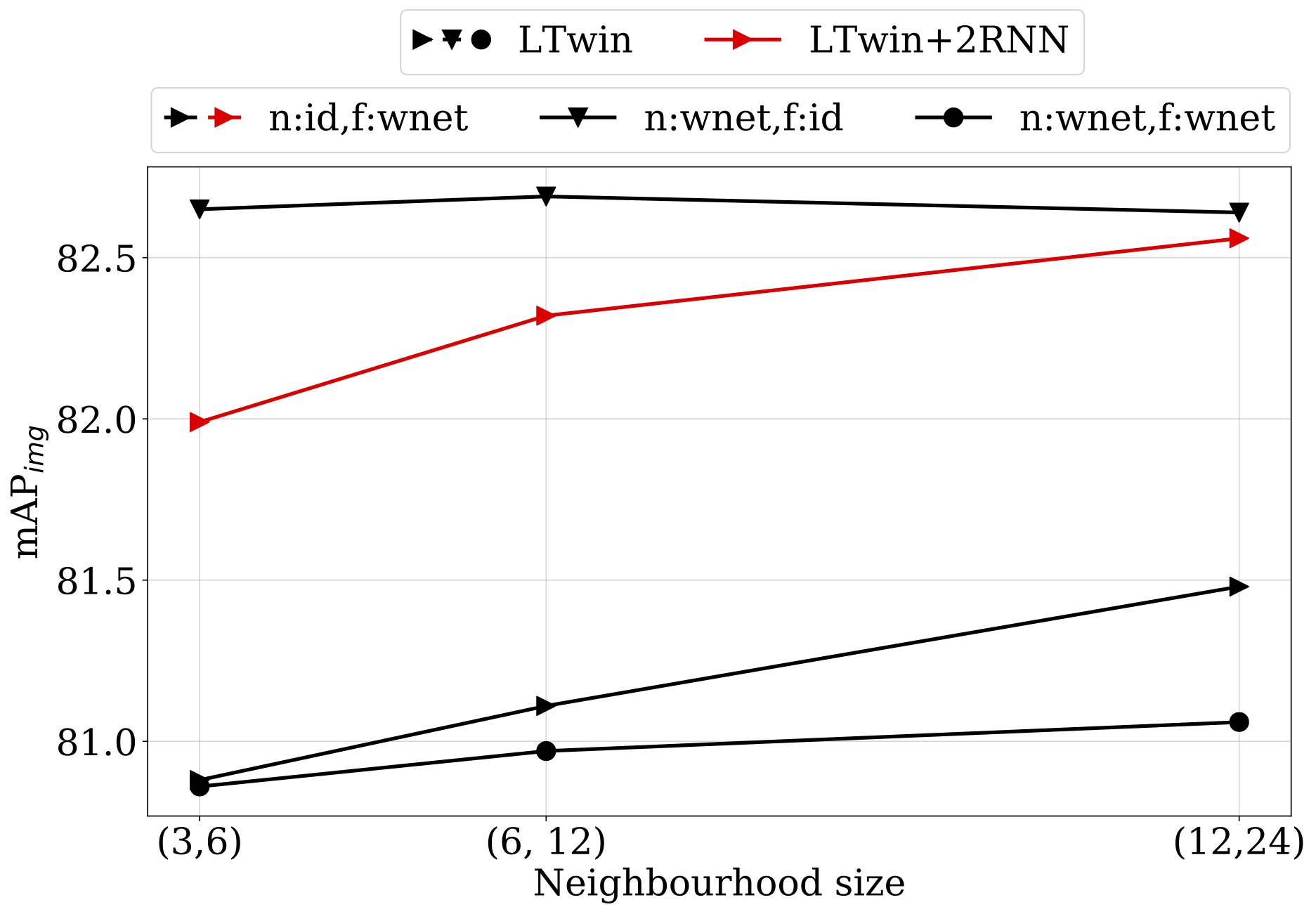}\\
\caption{mAP$_{lab}$ and mAP$_{img}$ for joint models varying the neighborhood size and considering $\pi=$\texttt{w2v} ($1^{st}$ row) and $\pi=$\texttt{wnet} ($2^{nd}$ row). Only relevant models and embedding combinations are reported. \texttt{n} refers to the embedding used for neighbors retrieval while \texttt{f} to embedding used to metadata representation.}
\label{fig:AllJoint}
\end{figure}

\subsubsection{Joint Models}
We firstly analyze the \textit{naive} case, i.e., $\pi = $ \texttt{id} (Figure~\ref{fig:AllJointid}) and then introduce semantic mapping (Figure~\ref{fig:AllJoint}). 
The simplest and worst-performing model is \texttt{LTN+Vecs} fed with
raw binary vectors; it shows quasi-linear improvement
w.r.t. neighborhood. \texttt{LZIP}, which uses a RNN, improves
uniformly upon it and achieves very good mAP$_{lab}$ and mAP$_{img}$
from the start but tends to exhibit a mild decrease in performance
with neighborhood size, along with \texttt{LTwin+2RNN}. In turn,
\texttt{LTwin} achieves good mAP$_{img}$ but comparatively poor
mAP$_{lab}$; \texttt{LTwin+RNN} achieves roughly comparable
performance, but shows linear improvement with $m$. \texttt{LZIP}, at
small $(m,M)$, and \texttt{LTwin+2RNN} are the best-performing
models showing that early fusion and RNNs are beneficial to increase network performance, with \texttt{LTwin} comfortably in the middle for mAP$_{img}$. Unfortunately, \texttt{LZIP} and \texttt{LTwin+2RNN} are also by far the longest to train by an order of magnitude (we just need to consider the breadth of the unrolled graph for non-trivial neighborhood sizes).

The addition of semantic metadata transforms can give a significant boost to performance, in addition to the benefits w.r.t. robustness of the model to vocabulary changes and applicability to a different database than the one used for training. The performance of all architectures is boosted when they are fed transformations computed from \texttt{word2vec} vectors through Eq.~\ref{eq:weight} instead of plain binary vectors. 
All models tend to saturate around (mAP$_{lab}$, mAP$_{img}$) = $(.63, .83)$. This appears to be the case for \texttt{LZIP}, even without any sort of $\pi$. It may be the case that the simpler \texttt{LTwin} can match the performance of the more complex models once provided with \texttt{word2vec} mappings. \texttt{LTwin} (f: \texttt{word2vec}) performs as well as \texttt{LTwin} (n: \texttt{word2vec}, f: \texttt{word2vec}), or even better; the same goes for its \texttt{LZip} siblings (by a considerably minor margin). We speculate that the ability of the network to learn to take maximal advantage of semantic embeddings overshadows the effect of their use in neighborhood generation and using \texttt{word2vec} vectors in the neighborhood generation process might therefore be unnecessary.  \texttt{LTwin} (f: \texttt{word2vec}) emerges as the superior model. As expected, \texttt{WordNet} results in poor performance. Notice also how \texttt{LTwin} (feed: \texttt{WordNet}) is particularly sensitive to neighborhood size.

\section{Conclusion}
\label{sec:conclusion}
We have shown that common visual models to classify images, based on metadata to retrieve neighbors, can be improved considering semantic mappings and recurrent neural networks. We have characterized the performance of a variety of visual and joint models and their variability. Our models outperform for several metrics state-of-the-art approaches. We have also shown that semantic mappings can be highly effective in improving performance, besides achieving robustness to changes in metadata vocabulary and quality of neighborhoods.

\bigskip
\paragraph*{Acknowledgements}
We gratefully acknowledge the support of NVIDIA for their donation of GPUs used in this research.
We also acknowledge the UNIPD CAPRI Consortium, for its support and access to computing resources.






\bibliographystyle{IEEEtran}
\bibliography{IEEEabrv,icpr20}

\end{document}